\documentclass[10pt,twocolumn,letterpaper]{article}

\usepackage[algorithms]{wacv}      

\usepackage{graphicx}
\usepackage{amsmath}
\usepackage{amssymb}
\usepackage{booktabs}
\usepackage{multirow}
\usepackage[x11names,dvipsnames,table]{xcolor} 
\usepackage{wrapfig}
\usepackage[normalem]{ulem}
\usepackage{balance}

\usepackage[pagebackref,breaklinks,colorlinks]{hyperref}

\usepackage[capitalize]{cleveref}
\crefname{section}{Sec.}{Secs.}
\Crefname{section}{Section}{Sections}
\Crefname{table}{Table}{Tables}
\crefname{table}{Tab.}{Tabs.}

\def\confYear{2025}

\newcommand{\subsubfour}[1]{\vspace*{1mm}{\noindent\bf #1}}
\newcommand{\subsubsubfour}[1]{\vspace*{0.5mm}{\noindent\bf #1}}

\begin{document}


\title{SimpsonsVQA: Enhancing Inquiry-Based Learning with a Tailored Dataset}

\author{Ngoc Dung Huynh\\
Deakin University\\
Australia\\
{\tt\small ndhuynh@deakin.edu.au}
\and
Mohamed Reda Bouadjenek\\
Deakin University\\
Australia\\
{\tt\small reda.bouadjenek@deakin.edu.au}
\and
Sunil Aryal\\
Deakin University\\
Australia\\
{\tt\small sunil.aryal@deakin.edu.au}
\and
Imran Razzak\\
The University of New South Wales\\
Australia\\
{\tt\small imran.razzak@unsw.edu.au}
\and
Hakim Hacid\\
Technology Innovation Institute\\
United Arab Emirates\\
{\tt\small hakim.hacid@tii.ae}
}

\maketitle

\begin{abstract}
\vspace{-0.5cm}

Visual Question Answering (VQA) has emerged as a promising area of research to develop AI-based systems for enabling interactive and immersive learning.  
Numerous VQA datasets have been introduced to facilitate various tasks, such as answering questions or identifying unanswerable ones.
\textcolor{black}{However, most of these datasets are constructed using real-world images, leaving the performance of existing models on cartoon images largely unexplored.}
Hence, in this paper, we present ``SimpsonsVQA'', a novel dataset for VQA derived from The Simpsons TV show, designed to promote inquiry-based learning.
Our dataset is specifically designed to address not only the traditional VQA task but also to identify irrelevant questions related to images, as well as the reverse scenario where a user provides an answer to a question that the system must evaluate (e.g., as correct, incorrect, or ambiguous).
It aims to cater to various visual applications, harnessing the visual content of ``The Simpsons'' to create engaging and informative interactive systems. 
SimpsonsVQA contains approximately 23K images, 166K QA pairs, and 500K judgments  (\url{https://simpsonsvqa.org}).
\textcolor{black}{Our experiments show that current large vision-language models like ChatGPT4o  underperform in zero-shot settings across all three tasks, highlighting the dataset's value for improving model performance on cartoon images. }
We anticipate that SimpsonsVQA will inspire further research, innovation, and advancements in inquiry-based learning VQA.

\end{abstract}

\vspace{-0.75cm}

\section{Introduction}
\label{sec:intro}

Visual Question Answering (VQA) is a promising research field that lies at the intersection of Computer Vision (CV) and Natural Language Processing (NLP) to enable machines to answer questions about visual content~\cite{Antol_2015_ICCV,ren2015exploring,Zhang_2016_CVPR,ma2016learning,goyal2017making}. 
The research interest in VQA has encouraged the creation of numerous datasets for constructing and evaluating VQA models including VQA v1.0~\cite{Antol_2015_ICCV}, VQA v2.0  \cite{goyal2017making}, and GQA~\cite{hudson2019gqa}. 
In addition, many datasets are purposefully crafted for specialized applications in practical domains such as healthcare~\cite{hasan2018overview, Lau2018ADO,abacha2020overview}, diagnosing medical images~\cite{lin2021medical,10.1145/3460426.3463584}, cultural heritage \cite{sheng2016dataset, garcia2020dataset}, aiding customer service~\cite{8269806}, enhancing entertainment experiences~\cite{Gordon_2018_CVPR}, and generating captions for social media content~\cite{Vinyals_2015_CVPR}.

\begin{figure}[t]

    \subfloat[]{\includegraphics[width=.45\linewidth]{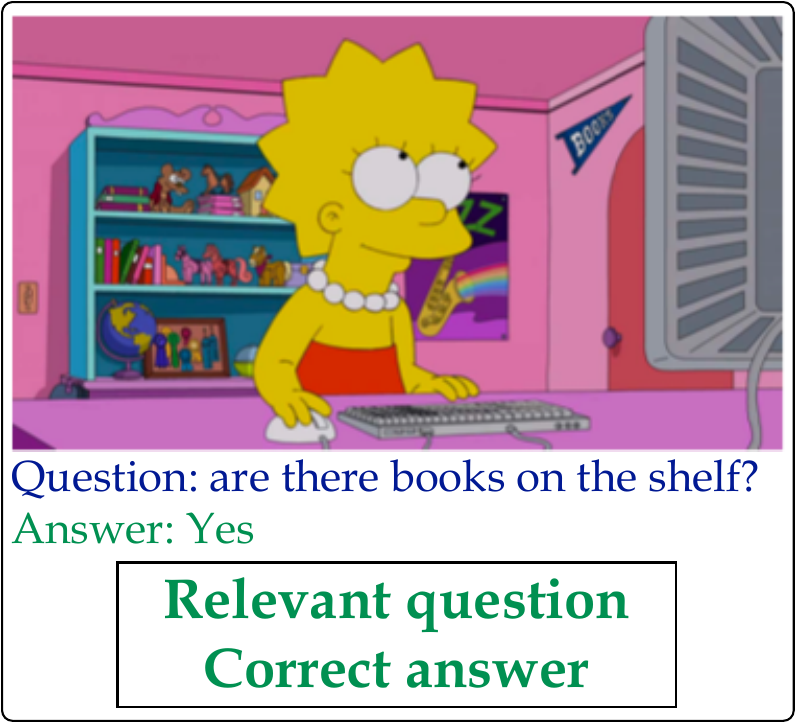}      \label{fig:SimpsonsVQA_examples1}
    }
    \subfloat[]{\includegraphics[width=.45\linewidth]{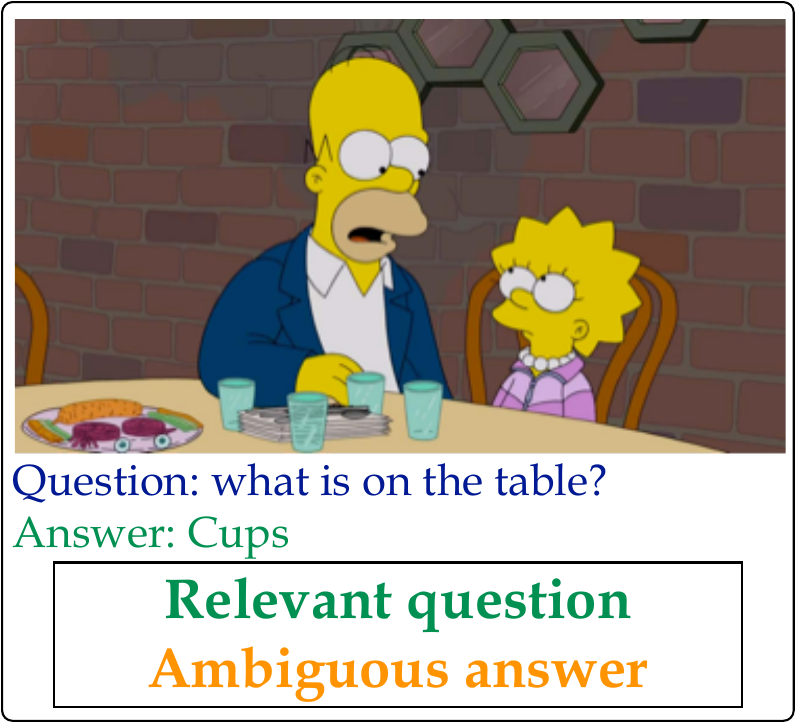}      \label{fig:SimpsonsVQA_examples2}
    }\\
     \subfloat[]{\includegraphics[width=.45\linewidth]{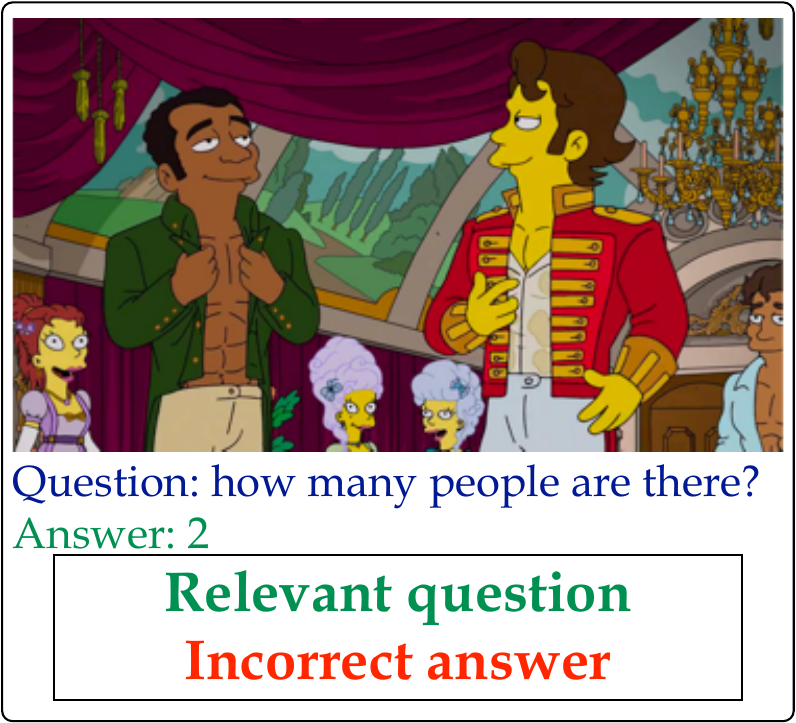}      \label{fig:SimpsonsVQA_examples3}
    }
    \subfloat[]{\includegraphics[width=.45\linewidth]{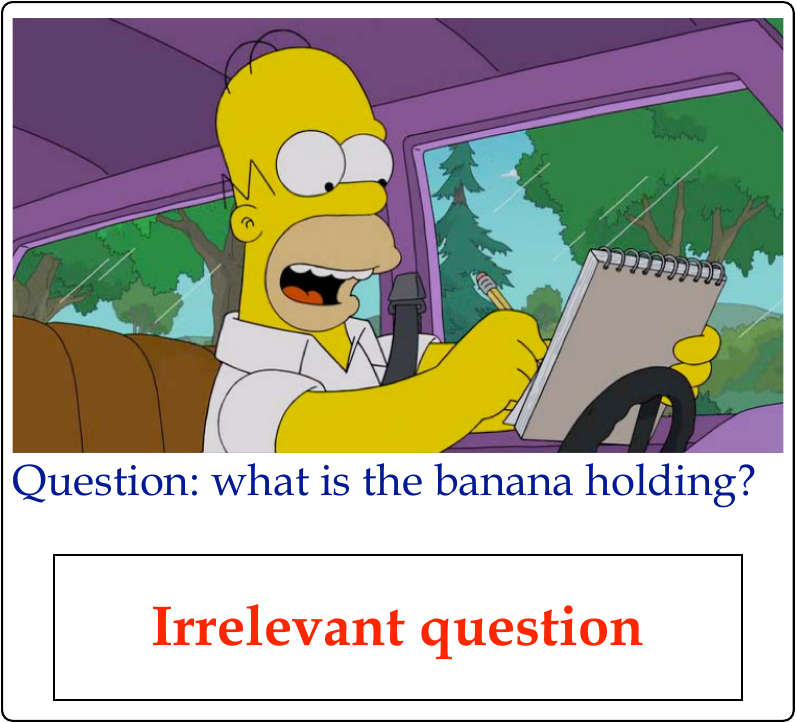}        \label{fig:SimpsonsVQA_examples4}
    }

    \vspace{-0.3cm}
    \caption{Examples from the SimpsonsVQA dataset.}
    \label{fig:SimpsonsVQA_examples}
    \vspace{-0.75cm}

\end{figure}

Despite the keen interest in VQA, the majority of the aforementioned datasets are primarily designed for \textbf{Scenario (1)}, where individuals ask \textit{relevant} questions about the content of an image, often with the aim of aiding visually impaired people~\cite{Antol_2015_ICCV, goyal2017making}. 
Conversely, several datasets have emerged to tackle \textbf{Scenario (2)}, involving individuals posing \textit{irrelevant} questions~\cite{ray2016question, stengel2022did, bhattacharya2019does, rajpurkar2018know, van2023document, 8578478, toor2017question, chandrasekaran2018explanations, mashrur2023robust, liu2018inverse, mahendru2017promise}. 
These questions should be intentionally left unanswered to prevent confusion and promote trust.
We argue in this paper that the existing literature has overlooked \textbf{Scenario (3)}, where an individual provides an answer to a question related to an image, requiring the system to evaluate it, e.g., ``Correct'', ``Incorrect'', or ``Ambiguous''.
These three scenarios are particularly relevant for individuals with cognitive impairments  and within educational contexts, especially early-age education. 
In these settings, individuals with developmental disorders may not only ask coherent questions but also pose irrelevant or inconsistent ones and provide incorrect answers related to visual content.

In this paper, we present ``SimpsonsVQA'', a \textit{unified} VQA dataset that can be used to address the three scenarios described above, fostering the development of intelligent systems that promote inquiry-based 
learning for individuals with cognitive disabilities and within early-age education.
\textcolor{black}{Unlike traditional datasets that primarily feature photorealistic images, our dataset leverages the cartoon imagery from \textit{The Simpsons} TV show, chosen for its , cultural relevance, and the ability to explore domain gaps between cartoon and real-world images.
Also, we argue that scenery, cartoons, and sketches are integral parts of our lives and should be considered in developing AI applications.
Additionally, the dataset  provides a valuable opportunity to assess the robustness of large vision-language models, such as Llava \cite{liu2024llavanext}, which are predominantly trained on photorealistic datasets. }

SimpsonsVQA incorporates triples consisting of images, questions, and answers, with evaluating judgments. 
For example, in Figure~\ref{fig:SimpsonsVQA_examples}, we show a sample of images along with free-form, open-ended, natural-language questions about the images, as well as their corresponding natural-language answers.
For instance, the image in Figure~\ref{fig:SimpsonsVQA_examples1} is linked to a relevant question and a correct answer. 
The image in Figure~\ref{fig:SimpsonsVQA_examples2} has a relevant question and an ambiguous or partially correct answer. 
Meanwhile, the image in Figure~\ref{fig:SimpsonsVQA_examples3} is associated with a relevant question but an incorrect answer, and the image in Figure~\ref{fig:SimpsonsVQA_examples4} is paired with an irrelevant question.
In total, SimpsonsVQA contains approximately 23K images, 166K QA pairs, and approximately 500K judgments.

We summarize the key contributions of this work as follows: 
(i) \textit{we introduce a new VQA task}, specifically focused on assessing candidate answers;
(ii) \textit{we present ``SimpsonsVQA''}, a \textit{unified} VQA dataset that is explicitly tailored to address the aforementioned tasks to foster inquiry-based learning;
and (iii) \textit{we conduct a comprehensive evaluation} with the aim to benchmark the SimpsonsVQA dataset using  various state-of-the-art VQA models.
SimpsonsVQA has real-world applications in enhancing educational tools and assistive technologies by enabling interactive, inquiry-based learning through AI systems that can handle diverse visual questions, assess relevance, and evaluate answer correctness, thereby supporting early-age education and individuals with cognitive impairments.
\textcolor{black}{Besides, our findings indicate that while models trained on the \textit{SimpsonsVQA} dataset perform well on our cartoon-based test set, advanced models like Llava \cite{liu2024improved} and ChatGPT4o \cite{chatgpt4o}, which typically excel in various zero-shot VQA challenges, encounter difficulties when dealing with cartoon images.}

\section{Related Work}
\label{sec:RelatedWork}

\subsubfour{Existing datasets:}
Several VQA datasets have been introduced for research purposes, as summarized in Table~\ref{tbl:datasets}. 
The VQA v1.0 dataset~\cite{Antol_2015_ICCV} is often credited with popularizing the VQA task. 
It was introduced as one of the first large-scale VQA datasets, containing real-world images paired with open-ended questions, and it has been widely used to develop and benchmark VQA models. 
Later, it was expanded and improved in VQA v2.0~\cite{goyal2017making} to address language bias, establishing itself as the benchmark dataset for the VQA task. 
Since then, various datasets with diverse objectives have been released, such as DAQUAR~\cite{malinowski2014multi} that focuses on indoor scenes, 
Visual Genome~\cite{krishna2017visual} and Visual7W~\cite{zhu2016visual7w}  for information about the relationships between objects,
CLEVR~\cite{johnson2017clevr} in which images are rendered geometric shapes, 
and GQA \cite{hudson2019gqa} for visual reasoning and compositional answering.
As shown in Table~\ref{tbl:datasets}, some datasets have images, questions and/or answers generated and/or selected entirely through manual processes, while others involved automated procedures~\cite{wang2017fvqa,ren2015exploring,malinowski2014multi,hasan2018overview}.
\textcolor{black}{While existing VQA datasets, such as VQA v2.0 and GQA, have advanced the field by providing large-scale benchmarks for evaluating model performance on photorealistic images, they often lack the complexity introduced by non-photorealistic like those in SimpsonsVQA.
}.

\begin{table}[t]
\centering

\caption{Popular VQA datasets. 
VIP - Visual Impired People; 
AG - Answer Grounding;
OE: Open Ended;
MC: Multi-Choice.}
\label{tbl:datasets}
\vspace{-0.2cm}
\rowcolors{2}{gray!15}{white}
\resizebox{\linewidth}{!}{
\begin{tabular}{|c |l |c |c|c |c |c |c|} 
\hline
\rowcolor{gray!50}
&\textbf{Name} & \textbf{Year} & \textbf{Domain} &\textbf{\#Images} & \textbf{\#Questions} & \textbf{Type} &\textbf{Automated} \\
\hline

1&\href{https://arxiv.org/abs/1505.00468}{VQA v1.0} \cite{Antol_2015_ICCV} & 2015 & General &204,721 & 614,163  & OE\&MC & No\\
2&\href{https://arxiv.org/abs/1505.00468}{VQA v1.0} \cite{Antol_2015_ICCV}& 2015 & Abstract Scene  &50,000 & 150,000 & OE\&MC&No\\
3&\href{https://arxiv.org/abs/1505.02074}{COCO-QA} \cite{ren2015exploring}& 2015 & General &123,287 & 117,684 & OE& \textbf{Yes}\\
4&\href{https://arxiv.org/abs/1511.05099}{Binary-VQA} \cite{Zhang_2016_CVPR}& 2015 & Abstract Scene & 50,000 & 150,000 & MC& No\\
5&\href{https://proceedings.neurips.cc/paper/2015/hash/fb508ef074ee78a0e58c68be06d8a2eb-Abstract.html}{FM-IQA} \cite{gao2015you}& 2015 & General & 158,392 & 316,193 & OE& No\\
6&\href{https://arxiv.org/pdf/1511.02570.pdf}{KB-VQA} \cite{wang2015explicit}& 2015 & KB-VQA & 700 & 2,402 & OE&No\\
7&\href{https://arxiv.org/abs/1602.07332}{VG} \cite{krishna2017visual}& 2016 & General & 108,077 & 1,700,000  & OE&Yes\\
8&\href{https://arxiv.org/abs/1511.02799}{SHAPE} \cite{andreas2016neural}& 2016 & Abstract Shape  & 15,616 & 244 & MC&Yes\\
9&\href{https://www.semanticscholar.org/paper/A-Dataset-for-Multimodal-Question-Answering-in-the-Sheng-Gool/10be82098017fc2d60b0572cea8032afabad5d1a}{Art-VQA} \cite{sheng2016dataset}& 2016 & Cultural Heritage & 16 & 805 & OE& No\\
10&\href{https://arxiv.org/abs/1606.05433}{FVQA}  \cite{wang2017fvqa}  & 2017 & KB-VQA & 1,906 & 4,608 & OE& \textbf{Yes}\\
11&\href{https://arxiv.org/abs/1410.0210v4}{DAQUAR}  \cite{malinowski2014multi}  & 2017 & General & 1,449 & 12,468 & OE&\textbf{Yes}\\
12&\href{https://arxiv.org/abs/1511.03416}{Visual7W} \cite{zhu2016visual7w} & 2017 & General &47,300 & 327,939 &  MC& Yes\\
13&\href{https://arxiv.org/abs/1612.00837}{VQA v2.0} \cite{goyal2017making} & 2017 & General &200,000 & 1,100,000 & OE\&MC& No\\
14&\href{https://arxiv.org/abs/1612.06890}{CLEVR} \cite{johnson2017clevr}& 2017 &  Geometric Shapes &100,000 & 853,554 & OE& Yes\\
15&\href{https://arxiv.org/abs/1712.00377}{VQA-CP1} \cite{agrawal2018don}& 2017 & General & 205,000 & 370,000 & OE& No\\
16&\href{https://arxiv.org/abs/1712.00377}{VQA-CP2} \cite{agrawal2018don}& 2017 & General & 219,000 & 658,000 & OE& No\\
17&\href{https://openaccess.thecvf.com/content_cvpr_2017/papers/Hussain_Automatic_Understanding_of_CVPR_2017_paper.pdf}{AD-VQA} \cite{hussain2017automatic}& 2017 &  Advertisement & 64,832 & 202,090 & OE& No\\
18&\href{https://arxiv.org/pdf/1703.09684}{TDIUC} \cite{kafle2017analysis}& 2020 & General& 167,437 & 1,654,167 & OE& Yes\\
19&\href{https://ceur-ws.org/Vol-2125/paper_212.pdf}{VQA-MED-18} \cite{hasan2018overview}& 2018 & Medical & 2,866 & 6,413 & OE& \textbf{Yes}\\
20&\href{https://www.nature.com/articles/sdata2018251}{VQA-RAD} \cite{Lau2018ADO}& 2018 & Medical & 315 & 3,515 & OE& No\\
21&\href{https://vizwiz.org/tasks-and-datasets/vqa/}{VizWiz} \cite{8578478} & 2018 & VIP & 32,842 & 32,842& OE& No\\
22&\href{https://www.imageclef.org/2019/medical/vqa}{VQA-MED-19} \cite{abacha2019vqa}& 2019 &  Medical & 4,200 & 15,292 & OE& Yes\\
23&\href{https://arxiv.org/pdf/1904.08920v2.pdf}{TextVQA} \cite{singh2019towards}& 2019 & Text-VQA &28,408 & 45,336 & OE& No\\
24&\href{https://ieeexplore.ieee.org/document/8978122}{OCR-VQA} \cite{8978122}& 2019 & Text-VQA &207,572 & 1,002,146 & OE& \textbf{Yes}\\
25&\href{https://arxiv.org/pdf/2002.10215.pdf}{STE-VQA} \cite{wang2020general}& 2019 & Text-VQA & 21,047 & 23,887 & OE& No\\
26&\href{https://arxiv.org/abs/1905.13648}{ST-VQA} \cite{biten2019scene}& 2019 & Text-VQA & 22,020 & 30,471 & OE&No\\
27&\href{https://arxiv.org/pdf/1906.00067v2.pdf}{OK-VQA} \cite{marino2019ok}& 2019 & KB-VQA & 14,031 & 14,055 & OE& No\\
28&\href{https://arxiv.org/abs/1902.09506}{GQA} \cite{hudson2019gqa}& 2019 & General &113,000 & 22,000,000 & OE& \textbf{Yes}\\
29&\href{https://arxiv.org/pdf/1907.12861v1.pdf}{LEAF-QA} \cite{chaudhry2020leaf}& 2019 & FigureQA &240,000 & 2,000,000& OE&\textbf{Yes}\\
30&\href{https://arxiv.org/abs/2007.00398}{DOC-VQA} \cite{mathew2021docvqa}& 2020 & Text-VQA&12,767 & 50,000 & OE& No\\
31&\href{https://arxiv.org/abs/2008.12520}{AQUA} \cite{garcia2020dataset}& 2020 & Cultural Heritage & 21,383 & 32,345 & OE& \textbf{Yes}\\
32&\href{https://arxiv.org/abs/2003.07333}{RSVQA-low} \cite{8578478} & 2020 & Remote Sensor  & 772 & 77,232 & OE&\textbf{Yes}\\
33&\href{https://arxiv.org/abs/2003.07333}{RSVQA-high} \cite{8578478} & 2020 & Remote Sensor & 10,659 & 1,066,316 & OE&\textbf{Yes}\\
34&\href{https://ceur-ws.org/Vol-2696/paper_106.pdf}{VQA-MED-20} \cite{abacha2020overview}& 2020 &  Medical & 5,000 & 5,000 & OE&Yes\\
35&\href{https://aclanthology.org/2020.bionlp-1.6.pdf}{RadVisDial} \cite{kovaleva2020towards}& 2020 & Medical & 91,060 & 455,300& OE& Yes\\
36&\href{https://arxiv.org/abs/2003.10286}{PathVQA} \cite{he2020pathvqa}& 2020 & Medical& 4,998 & 32,799 & OE& Yes\\

37&\href{https://arodes.hes-so.ch/record/9062}{VQA-MED-21} \cite{ImageCLEF-VQA-Med2021}& 2021 & Medical &  5,500 &  5,500 & OE&Yes\\
38&\href{https://arxiv.org/abs/2102.09542}{SLAKE} \cite{liu2021slake}& 2021 & Medical & 642 & 14,000 & OE&No\\
39&\href{https://arxiv.org/abs/2105.14517}{GeoQA} \cite{chen2021geoqa}& 2021 & Geometry Problems & 5,010 & 5,010 & MC&No\\
40&\href{https://arxiv.org/abs/2101.11272}{VisualMRC} \cite{tanaka2021visualmrc}& 2021 & Text-VQA & 10,197 & 30,562 & OE& Yes\\
41&\href{https://opendatalab.com/A-OKVQA}{A-OKVQA } \cite{schwenk2022okvqa}& 2022 &   KB-VQA & 23,700 & 37,687 & OE& No\\
42&\href{https://arxiv.org/abs/2202.01993}{VizWiz-Ground} \cite{chen2022grounding}& 2022 & VIP + AG & 9,998 & 9,998 & -& Yes\\
43&\href{https://ceur-ws.org/Vol-3357/invited1.pdf}{WSDM Cup} \cite{TolokaWSDMCup2023}& 2023 & AG & 45,119 & 45,119 & -& No\\
\hline
\textbf{44}&\textbf{SimpsonsVQA}& \textbf{2024} & \textbf{Cartoon} & \textbf{23,269} & \textbf{103,738 }& \textbf{OE} & \textbf{Yes}\\

\hline
\end{tabular}
}
\vspace{-0.5cm}

\end{table}

\subsubfour{Question relevance:} The common belief when gathering responses to visual questions is that questions can be answered using the provided image~\cite{Antol_2015_ICCV, andreas2016neural, gao2015you, goyal2017making, johnson2017clevr, krishna2017visual, malinowski2014multi, ren2015exploring, wang2015explicit, wang2017fvqa}. 
However, in practice, not everyone asks questions directly related to the visual content~\cite{davis2020unanswerable}, especially early-age learners.
In VQA v1.0~\cite{Antol_2015_ICCV}, Ray et al.~\cite{ray2016question} conducted a study where they randomly selected 10,793 question-image pairs from a pool of 1,500 unique images. Their findings revealed that 79\% of the questions were unrelated to the corresponding images.
Hence, a VQA system should avoid answering an irrelevant question to an image,  as doing so may lead to considerable confusion and a lack of trust.
The exploration of question relevance has been extensively explored in the literature, leading to the development of numerous methods and algorithms aimed at avoiding answering irrelevant questions. 
Notable contributions include works such as \cite{ray2016question, stengel2022did, bhattacharya2019does, rajpurkar2018know, van2023document, 8578478, toor2017question, chandrasekaran2018explanations, mashrur2023robust,mahendru2017promise}.
\textcolor{black}{
Our work with the \textit{SimpsonsVQA} dataset expands on previous efforts by integrating both relevant and irrelevant questions. Testing models with irrelevant queries enhances their ability to distinguish between relevant and unrelated inputs, improving reliability and ensuring accurate responses in real-world applications like educational and assistive technologies.
}

\subsubfour{Answer Correctness:}
\textcolor{black}{
While recent studies on VQA responses focuses on abstaining when uncertain to handle negative pairs and improve reliability \cite{whitehead2022reliable}, our method directly evaluates answers without abstaining, ensuring scalability and flexibility in real-world applications like education, assistive technologies, and customer services. The most relevant work, LAVE \cite{manas2023improving}, uses an LLM to evaluate answers based on reference alignment and context from the question and image. However, LAVE's reliance on reference answers limits independent evaluation. Our approach directly evaluates answers with image-based questions, eliminating the need for reference answers, enabling scalable, flexible, and autonomous VQA systems.
}

\vspace{-0.2cm}

\section{SimpsonsVQA Dataset}
We provide in the following an overview of the dataset. 
\subsection{Dataset Creation}
Due to the constraints imposed by limited time and budget, we adopted a pragmatic approach of automation to streamline the dataset construction process. 
In fact, many datasets listed in Table~\ref{tbl:datasets} have been created through partial automation methods~\cite{krishna2017visual,wang2017fvqa,ren2015exploring,malinowski2014multi,hasan2018overview,andreas2016neural,zhu2016visual7w,johnson2017clevr,abacha2019vqa,8978122}. 
To accomplish this, we employed a three-step approach: 
(1) harnessing the capabilities of Machine Learning models, particularly captioning models, to extract descriptions for each image; 
(2) employing ChatGPT to generate a diverse set of question-answer pairs using the obtained descriptions; 
and ultimately, (3) conducting a meticulous manual review by qualified workers on the AMT platform to judgments of accuracy and reliability.
In the following subsections, we provide a detailed description of these steps.

\subsubfour{Image Collection:}
We have collected cartoon images from the popular American sitcom, ``The Simpsons''. 
Focusing on seasons 24 to 33, which include 220 episodes and approximately 80 hours of content, we used an automated process to capture images every 5 seconds, initially gathering around 43,000 images. 
After manually filtering out about 1,200 inappropriate images (containing violence, weapons, or sexual content) and excluding images lacking substantial content, we employed the $k$NN algorithm~\cite{1053964} (with $k=3$) to remove duplicates.
This process resulted in a final dataset of 23,269 images.

\subsubfour{Image Captioning:} 
Image Captioning \cite{Vinyals_2015_CVPR}  combines CV and NLP to generate descriptive captions for images. 
\textcolor{black}{
We used the captioning OFA model~\cite{wang2022ofa}, originally trained on COCO\cite{lin2014microsoft}, which typically generates short captions focused on basic object descriptions.
To enable the generation of richer, longer captions capable of supporting diverse question-answer pairs per image, we fine-tuned OFA using the Localized Narratives ~\cite{PontTuset_eccv2020} and Image Paragraph Captioning datasets ~\cite{krause2016paragraphs}. These datasets enhanced the coherence and richness of multi-sentence descriptions.
Notably, it might not have captured character names or specific nuances of the TV show.
Since the questions were generated from the visual captions, they are primarily focused on visual elements. While the questions are generally clear, some present a greater level of difficulty. For instance, the question ``What is the object in the hand on the left-back wall?'' refers to a phone, which may not be immediately obvious without context or additional visual cues.
}

\subsubfour{Generating Question-Answer Pairs:} 
\textcolor{black}{We used ChatGPT ~\cite{openai2021}  to generate at least 10 question-answer pairs \textcolor{black}{alongside with question topics} for each image description, allowing the model to produce answers and topics freely to capture diverse responses.  Initially, we obtained over 1,000 unique answers. After manually removing inconsistent questions (e.g., those with predictable answers like "yellow" for skin color), we normalized the dataset by standardizing numerical answers (e.g., ``one'',``1'', to ``1''). To maintain simplicity and focus, we also dropped any answers that were longer than a single word or phrase, ensuring that the dataset remained concise and consistent. As a result, we finalized a dataset of 166,533 image-question-answer triples with around 200 unique answers.}

\begin{figure}[t]

    \subfloat[]{\includegraphics[width=.45\linewidth]{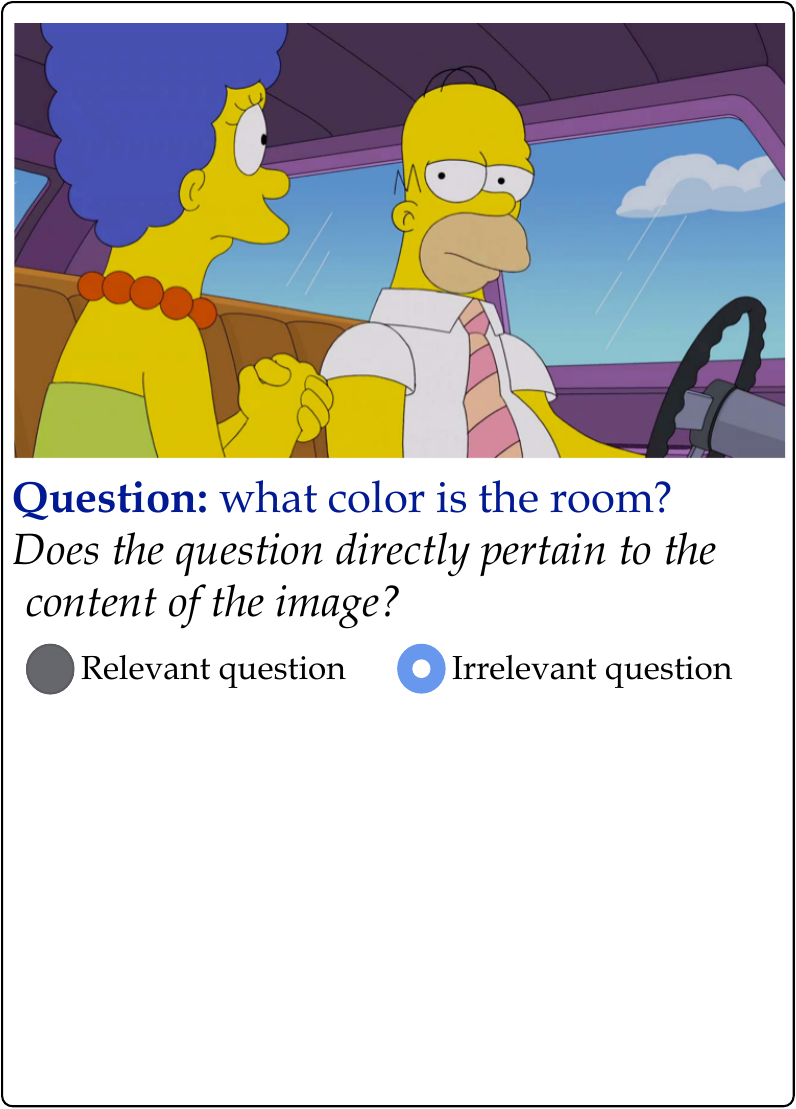}      \label{fig:AMT_Interface1}
    }
    \subfloat[]{\includegraphics[width=.45\linewidth]{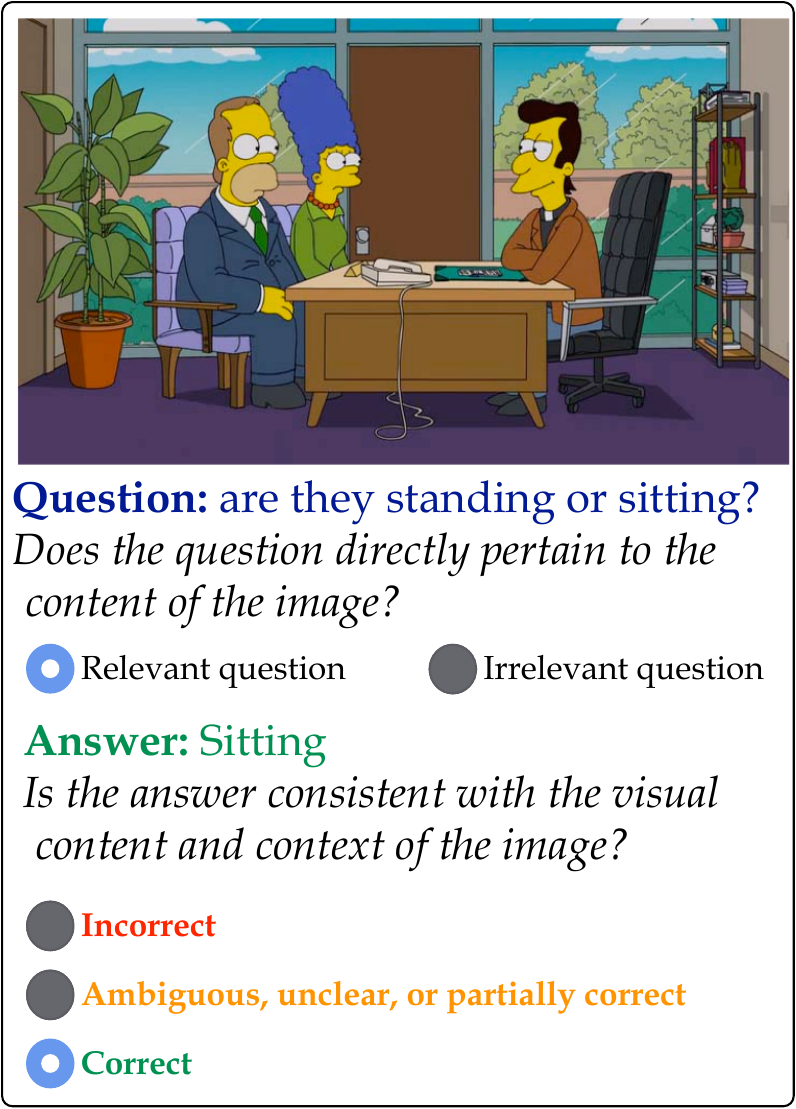}      \label{fig:AMT_Interface2}
    }
    \vspace{-0.3cm}
    \caption{AMT assessment interface.}
    \label{fig:AMT_Interface}
    \vspace{-0.5cm}

\end{figure}

\subsubfour{Assessing image-question-answer triples:} 
\textcolor{black}{
The above models are prone to errors, which can lead to the generation of irrelevant questions and/or incorrect answers. We acknowledge that the generated questions and answers may not reflect typical human learner errors, potentially creating a domain gap, which is important to consider for real-world applicability.
}
We employed the Amazon Mechanical Turk (AMT) platform to assess each image, question, and answer triple, using the interface depicted in Figure~\ref{fig:AMT_Interface}.
In particular, we engaged human evaluators through the AMT platform and tasked them with evaluating each triple according to particular criteria. 
Initially,
workers are
presented with an image and a question, and then they 
are
prompted to determine whether the question directly relates to the content of the image, offering a binary choice between ``\textit{relevant}'' or ``\textit{irrelevant}''.
If the worker chooses ``\textit{irrelevant}'', no additional action is needed for the given triple as in the case shown in Figure~\ref{fig:AMT_Interface1}.
Otherwise, as depicted in Figure~\ref{fig:AMT_Interface2}, the worker must evaluate the accuracy of the answer to the question and its alignment with the image context by selecting one of these options: 
(i) \textit{incorrect}, indicating that the provided answer is entirely wrong; 
(ii) \textit{ambiguous or partially correct}, suggesting that the answer is unclear, open to interpretation, or it includes some correct details but also incorporates incorrect or irrelevant elements, making its validity hard to determine; 
and (iii) \textit{correct}, implying that the answer is precise and directly addresses both the question and image.

To ensure the integrity of the evaluation process, each triple was assessed by three different workers. 
Rigorous eligibility criteria were enforced, allowing only individuals with a minimum approval rate of 99\% and a track record of at least 10,000 approved HITs (Human Intelligence Tasks) to participate in the evaluation of the triples.
To mitigate fraudulent or unreliable evaluations, each HIT required a minimum of 1 minute to be completed.

\subsection{Task Description}
\label{sec:Task_Description}

Considering a set of images $\mathcal{I}=\{\textbf{i}_{1}, \textbf{i}_{2}, \cdots\}$, a set of questions $\mathcal{Q}=\{\textbf{q}_{1}, \textbf{q}_{2}, \cdots\}$, a set of possible answers $\mathcal{A}=\{a_{1}, a_{2}, \cdots\}$, the SimpsonsVQA dataset is designed to emphasize three tasks as described below.

\subsubfour{Conventional VQA Task:}
Given a dataset $\mathcal{D} = \{(\textbf{i}^{(i)},\textbf{q}^{(i)},a^{(i)})\}_{i=1}^{m}$ with $m$ instances, the objective of this task is to develop a classification algorithm that learns a mapping function $f:(\mathcal{I},\mathcal{Q}) \rightarrow \mathcal{A}$, which associates each image-question pair $(\textbf{i}^{(i)},\textbf{q}^{(i)})$   with its corresponding correct answer $a^{(i)}$.
This initial task embodies the conventional VQA scenario, where an individual poses a question that the system is tasked to answer.

\subsubfour{Question relevance Task:}
Consider a dataset $\mathcal{D} = \{(\textbf{i}^{(i)},\textbf{q}^{(i)},y^{(i)})\}_{i=1}^{m}$, where  $y^{(i)}\in \{0,1\}$ represents a binary label indicating the relevance of a question to an image.
The objective of this task is to formulate a classification algorithm, denoted as $f : (\mathcal{I}, \mathcal{Q}) \rightarrow y$, aimed at learning a mapping that associates each image-question pair $(\mathbf{i}^{(i)}, \mathbf{q}^{(i)})$ with its corresponding binary label $y^{(i)}$.
In this scenario, an individual poses a question, and the system is required to assess its relevance to a provided image.

\subsubfour{Answer correctness Task:}
\textcolor{black}{Consider a dataset $\mathcal{D} = \{(\mathbf{i}^{(i)}, \mathbf{q}^{(i)}, a^{(i)}, z^{(i)})\}_{i=1}^{m}$, where $z^{(i)} \in \{\text{incorrect},$  
$ \text{ambiguous}, \text{correct}\}$ denotes a label signifying the alignment of an answer with an image-question pair.
The objective is to formulate a classification algorithm denoted as $f : (\mathcal{I}, \mathcal{Q}, \mathcal{A}) \rightarrow z$, aimed at learning a mapping that associates each image-question-answer triple $(\mathbf{i}^{(i)}, \mathbf{q}^{(i)}, a^{(i)})$ with its corresponding label $z^{(i)}$.}
This final task represents the scenario where an individual provides an answer to a question related to an image that the system must evaluate.

Overall, we believe that the three aforementioned scenarios hold significant relevance within the realm of assistive technology  applications. 
Our overarching goal is to craft interactive systems that are both captivating and enlightening, fostering and facilitating inquiry-based learning experiences.
\setlength{\columnsep}{10pt} 
\setlength{\intextsep}{3pt} 
\setlength{\abovecaptionskip}{5pt} 

\begin{wraptable}{r}{0.2\textwidth}
    \centering
        \caption{Dataset split.}
    \label{tab:dataset}
    \rowcolors{2}{gray!15}{white}

\resizebox{0.2\textwidth}{!}{%
    \begin{tabular}{|ccc|}
    \hline
    \rowcolor{gray!50}

     & \textbf{\#Image} & \textbf{\#QA pairs} \\\specialrule{0.1pt}{0pt}{0pt}
     Train & 13,961 & 115,663\\
     \hline
     Validation & 3,490 & 21,949\\
     \hline
     Test & 5,818 & 28,921\\
     \hline
     \hline
     Total & 23,269 & 166,533\\
     \hline

\end{tabular}
}
\end{wraptable}

\section{SimpsonsVQA Dataset Analysis}

In this section, we  analyze various aspects of the SimpsonsVQA dataset, including its characteristics and distribution patterns, while also providing insights obtained from analyzing its content.
As reported in Table \ref{tab:dataset}, the dataset is partitioned into three subsets: train, validation, and test.

It is important to note that, in order to uphold the integrity and confidentiality of the evaluation procedure, the test set remains both private and undisclosed.

\subsection{Question Analysis}

\begin{wrapfigure}{r}{0.18\textwidth}
    \centering
    \includegraphics[width=0.15\textwidth]{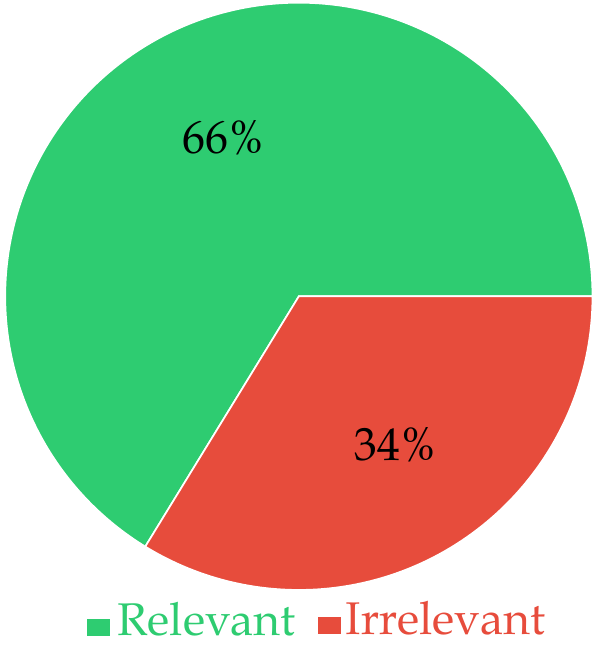}
    \caption{Ratio of question relevance as judged by AMT workers.}
    \label{fig:question_relevance}
\end{wrapfigure}
A total of 1,633 workers from AMT evaluated all the image-question-answer triples in our dataset.
As mentioned earlier, each triple has undergone evaluation by three distinct workers, each providing judgments on two aspects: 
\setlength{\columnsep}{10pt} 
\setlength{\intextsep}{3pt} 
\setlength{\abovecaptionskip}{2pt} 
(1) the question's relevance to the image content, and
(2) the accuracy of the answer in relation to the given image context. 
As illustrated in Figure~\ref{fig:question_relevance}, approximately 66\% of the questions generated by ChatGPT (totaling 80,137  questions) have been assessed as relevant by at least 2 workers for the corresponding images.
In contrast, only 34\% (35,526 questions) generated questions lack relevance to the images. 

\begin{figure}[h]
\centering
\includegraphics[width=1.07\linewidth]{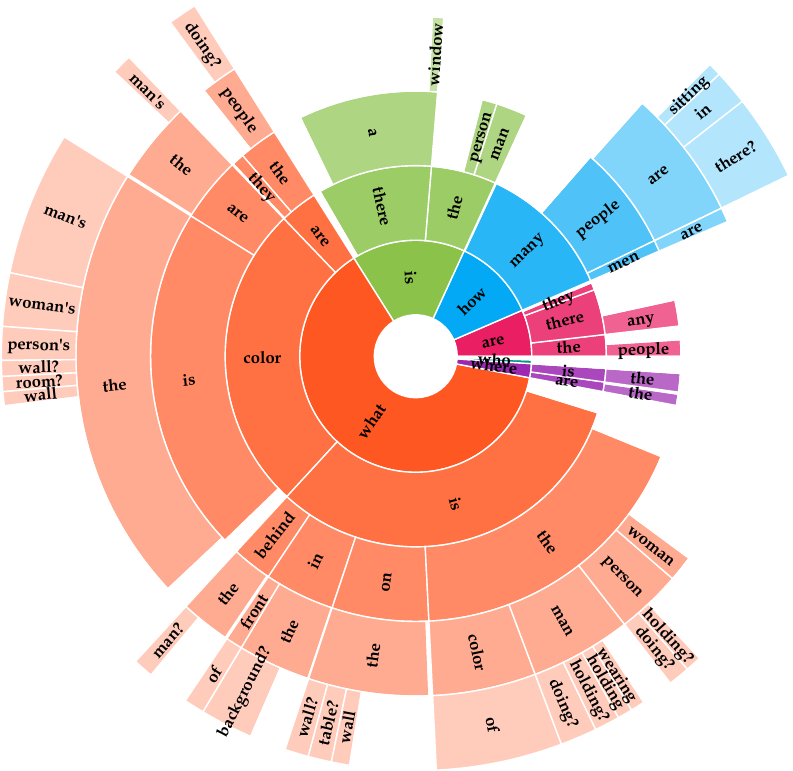}
\caption{Distribution of first words in questions. 
The words are arranged in an outward radiating pattern from the center, with their ordering based on frequency.
The size of the arcs corresponds to the number of questions containing each word.}
\label{fig:ques_dist}
\end{figure}

\subsubfour{Question Types:} 
\textcolor{black}{
Popular datasets like VQA v2.0 and VizWiz lack clearly defined question types, and while TDIUC \cite{kafle2017analysis} generates template-based questions, our dataset uses ChatGPT to create more diverse and nuanced questions, enhancing its richness, as shown in Figure \ref{fig:ques_dist}.
}
The majority of questions, approximately 55\% of the questions start with the word \textit{``what''}.
Following behind are questions beginning with \textit{``is''} and \textit{``how''}, which account for percentages ranging from 12\% to 20\%. 
Conversely, questions initiated by words like \textit{``are''}, \textit{``who''}, and \textit{``where''} make up a significantly smaller proportion. 
Furthermore, the most frequent question patterns include variations such as \textit{``what is the color...''}, \textit{``what color...''}, and \textit{``what is the man/woman/person doing/holding''}. 
Additionally, a substantial number of questions involve positional inquiries, such as \textit{``what is on/in/behind...''}, and there is also a significant presence of \textit{``how many...''} questions.

\setlength{\columnsep}{10pt} 
\setlength{\intextsep}{3pt} 
\setlength{\abovecaptionskip}{5pt} 

\begin{wrapfigure}{r}{0.2\textwidth}
    \centering
    \includegraphics[width=0.18\textwidth]{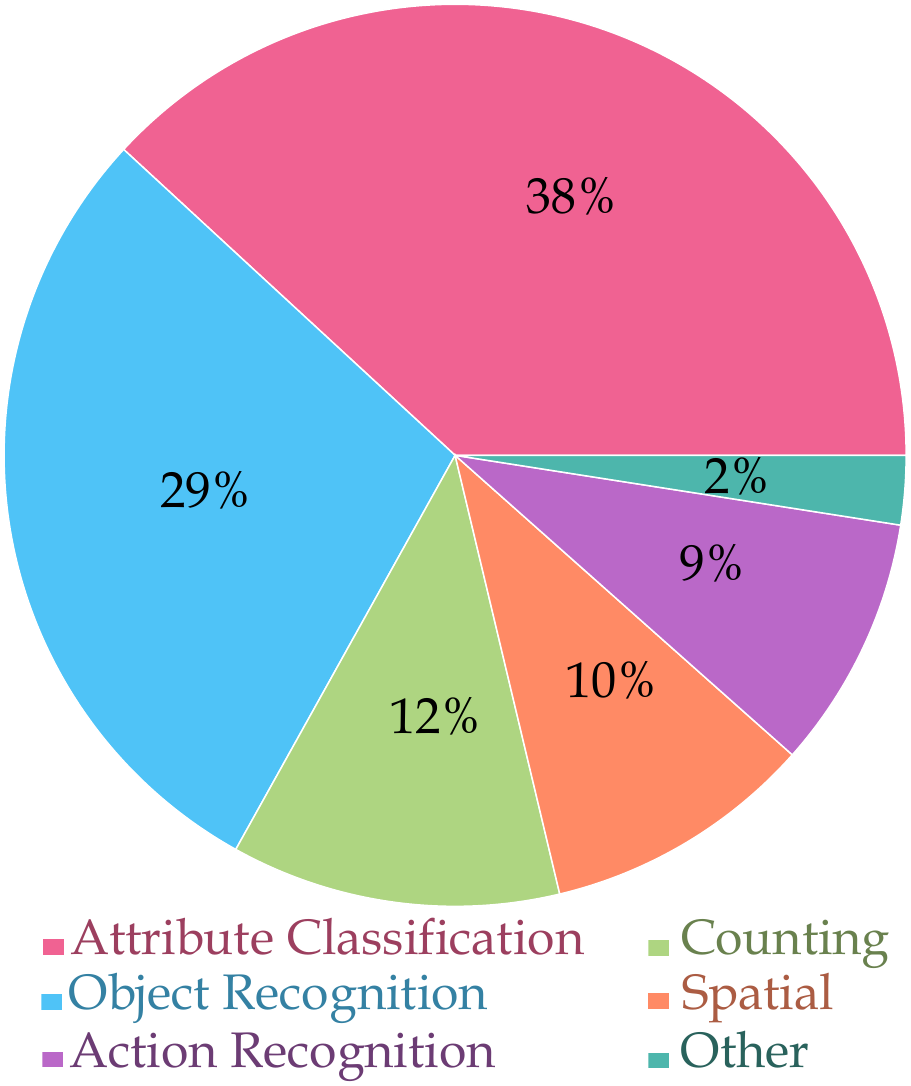}
    \caption{Question topics.}
    \label{tab:topic_pie}
\end{wrapfigure}

\begin{figure}[t]
    \centering   \includegraphics[width=1.03\linewidth]{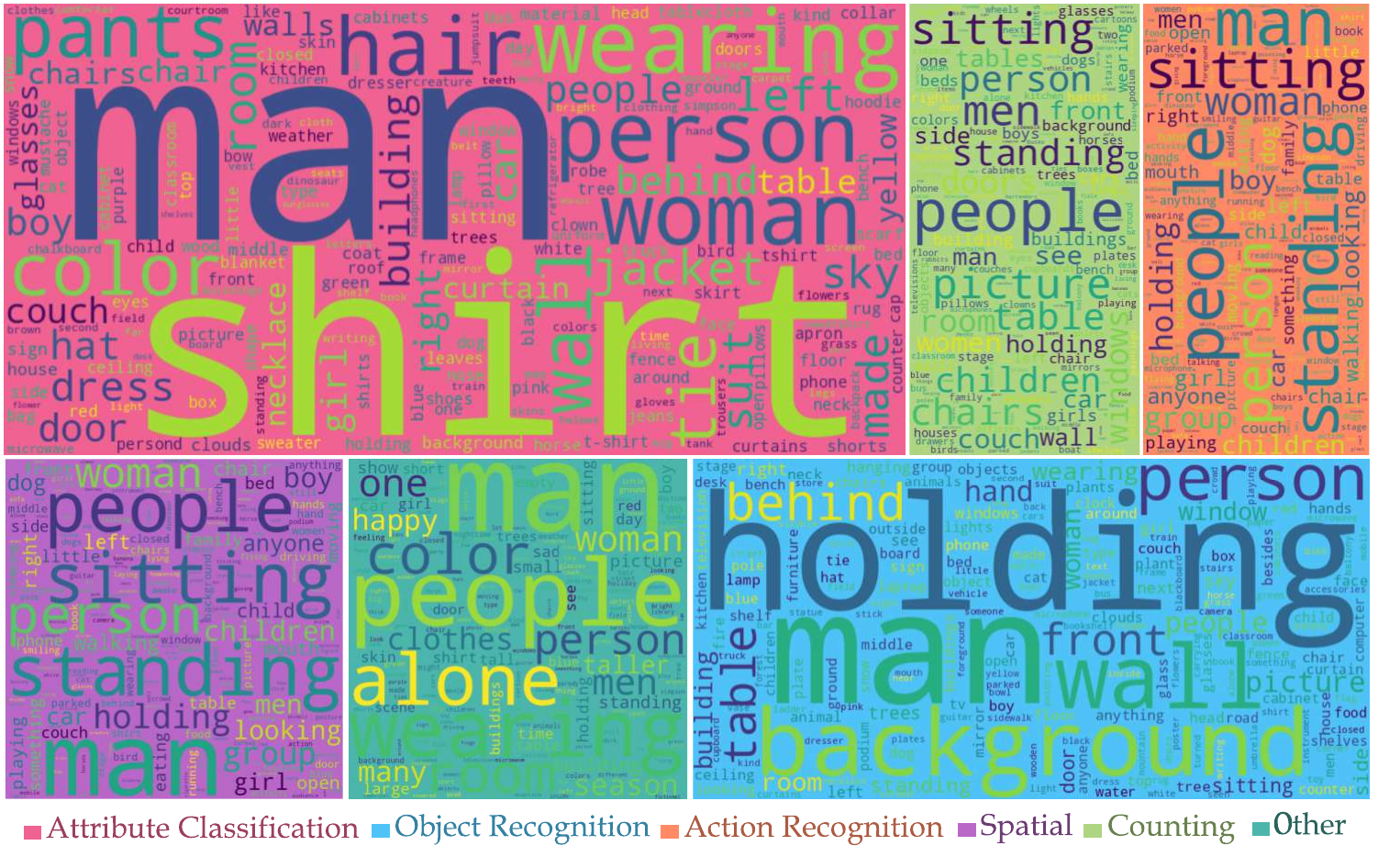}
    \caption{Word cloud of the most prominent terms in questions.}
    \label{fig:word_cloud_ques_type}
    \vspace{-0.5cm}
\end{figure}

\textbf{Question Topics:}
As shown in Figure~\ref{tab:topic_pie}, the questions cover a wide range of topics, encompassing attribute classification
38\%, object recognition 29\%,
counting 12\%, spatial reasoning 10\% , and
action recognition
9\%.
The remaining topics collectively represent a negligible percentage, totaling only about 2\% of the questions.
These diverse topics play a crucial role in fostering various developmental abilities for inquiry-based learning in early-age education. 
Figure~\ref{fig:word_cloud_ques_type} presents individual word cloud visualizations for each question type, capturing the distinctive vocabulary associated with different question categories. 
Each cloud highlights the frequency of specific terms, offering a visual insight into the unique linguistic characteristics of various question types.

\begin{figure}[h]
\centering
{\includegraphics[width=.99\linewidth]{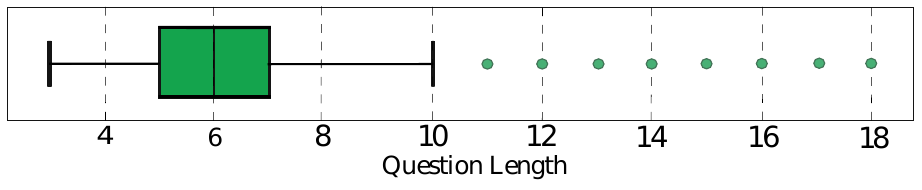}
}
\caption{Question length by number of words.}
\label{fig:question_len}
\end{figure}

\subsubfour{Question Length:} 
Figure \ref{fig:question_len} illustrates the distribution of question lengths, revealing that the majority of questions fall within the range of 3 (e.g., \textit{``are there balloons?''}) to 10 words, with a median of 6 words.
Notably, the longest question observed has a length of 18 words, \textcolor{black}{which is comparable to those found in other general VQA datasets.}

\subsection{Answer Analysis}
\setlength{\columnsep}{10pt} 
\setlength{\intextsep}{0pt} 
\setlength{\abovecaptionskip}{5pt} 
\begin{wrapfigure}{r}{0.2\textwidth}
\centering
\vspace{-0.4cm}
\includegraphics[width=0.2\textwidth]{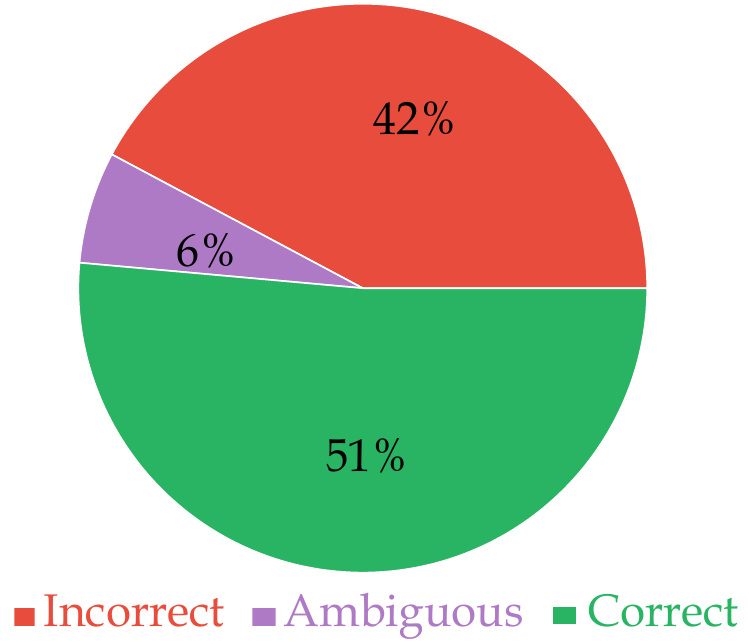}   
\caption{Worker Judgments on Triples.}
\label{fig:mturk_judgement}
\end{wrapfigure}

Figure \ref{fig:mturk_judgement} illustrates that approximately 51\% of the triples were assessed as \textit{``Correct''} by at least two workers, while roughly 42\% were deemed \textit{``Incorrect''} by at least two workers, and around 6\% were labeled as \textit{``Ambiguous''} by at least two workers.

\begin{figure}[h]
\captionsetup{justification=centering, singlelinecheck=false}
        \centering
        \includegraphics[width=0.98\linewidth]{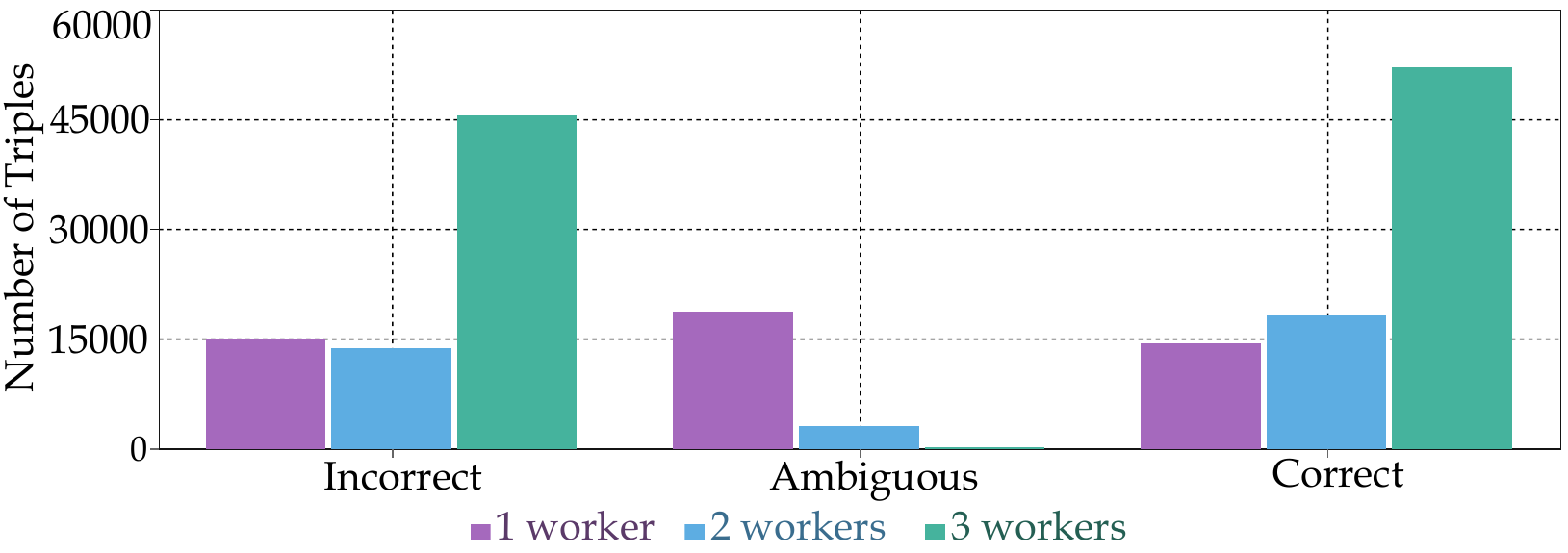}
        \caption{Assessment of triples by workers.}
        \label{fig:mturk_worker}
        \vspace{-0.15cm}
\end{figure}

In Figure \ref{fig:mturk_worker}, we notice that around 52,000 triples were judged \textit{``Correct''} by all three workers, and more than 18,000 triples were judged \textit{``Correct''} from exactly two workers. 
On the other hand, around 45,000 triples were unanimously judged as \textit{``Incorrect''} by all workers, and approximately 14,000 triples had agreement from exactly two workers labeling them as \textit{``Incorrect''}.
The number of triples judged partially ambiguous or partially correct was small.


\begin{figure}[h]
    \centering    
    \subfloat[30 most frequent answers.]{\includegraphics[width=0.72\linewidth]{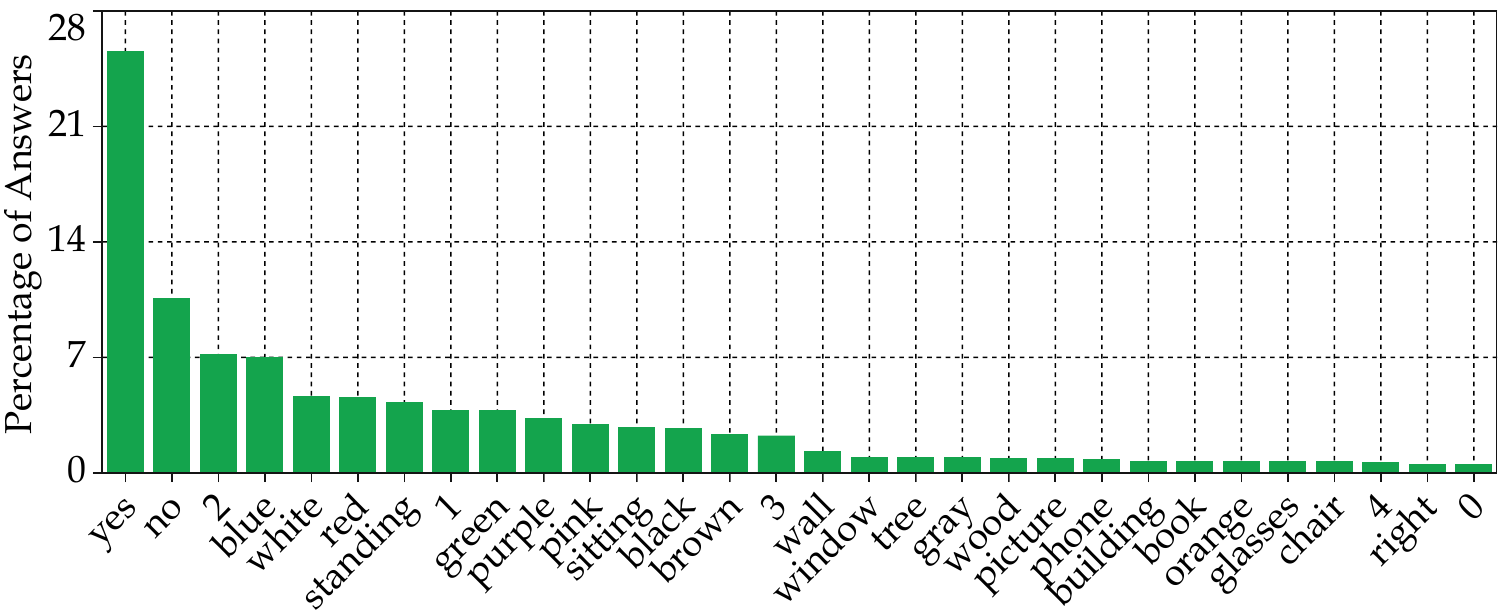}\label{fig:answers}}
   \subfloat[Answer types.]{ \includegraphics[width=0.28\linewidth]{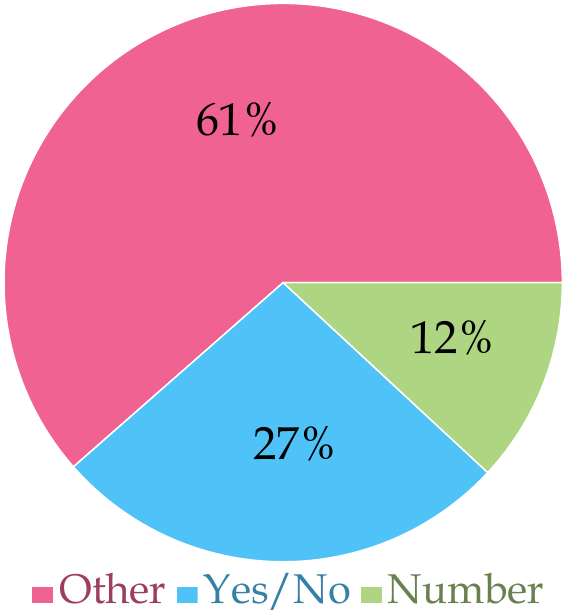}
   \label{fig:answer_type}}
    \caption{Answer distribution and statistics.}
    
\end{figure}

\subsubfour{Popular Answers:} All answers within the dataset consist of a single word.
Figure \ref{fig:answers} displays the top 30 answers with the highest frequency in the training set.
Notably, the answer ``yes'' predictably holds the top position, constituting 25\% of the answers, maintaining a notable lead of 11\% over the second-place answer, ``no''.
Among the 15 most frequent answers,
the majority tend to revolve around numbers or colors. 
This characteristic makes the dataset particularly suitable for educational applications.
Finally, as shown in Figure \ref{fig:answer_type}, 27\% of the questions prompted ``yes'' or ``no'' responses, while 12\% of the questions received numerical answers. The remaining 61\% of questions were answered in diverse ways.

\begin{figure*}[t]
     \centering
     \includegraphics[width=0.92\linewidth]{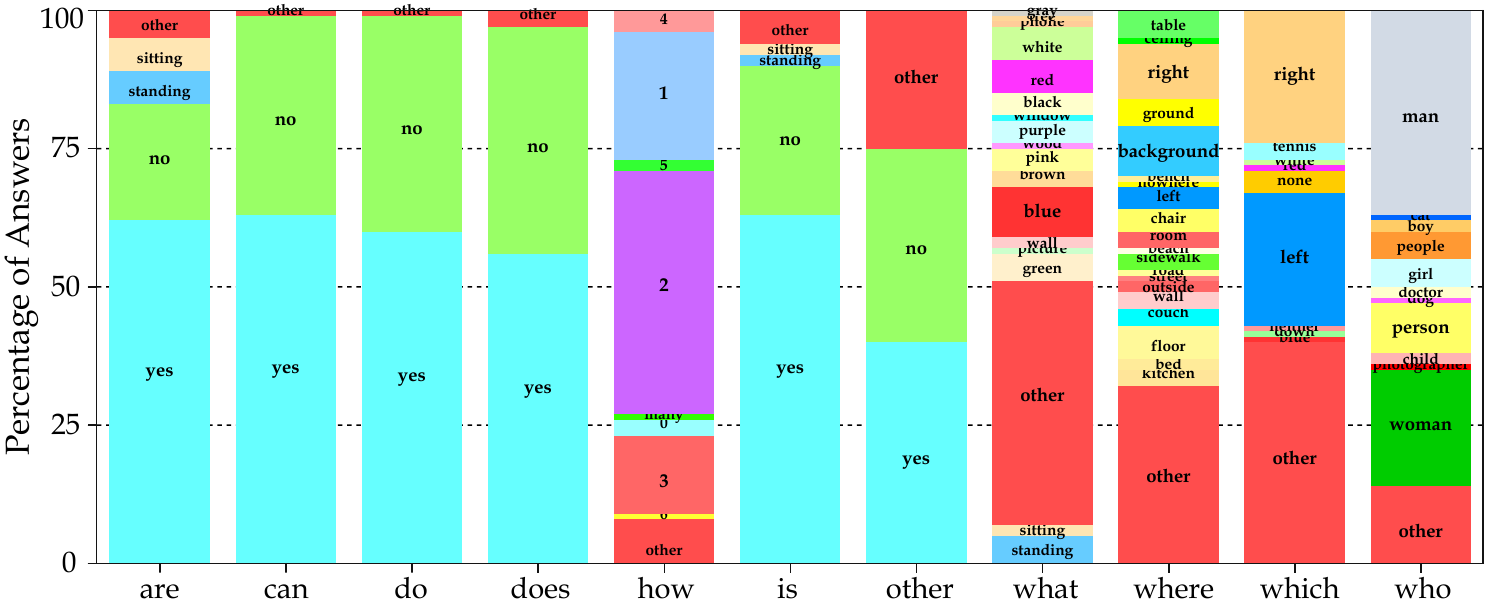}
     \vspace{-0.1cm}

     \caption{Answer distribution based on first words of questions.}     \label{fig:answer_first_word_question}
\vspace{-0.5cm}
 \end{figure*}

\subsubfour{Answers and Question Types:} Figure \ref{fig:answer_first_word_question} shows how different question types are answered. 
Questions starting with words like ``are'', ``can'', ``do'', ``does'', and ``is'' are mostly answered with either ``Yes/No''.
Surprisingly, questions beginning with ``How'' not only receive numeric answers but also frequently involve words like ``many'' and ``groups''. 
On the other hand, questions starting with ``what'', ``where'', and ``who'' have a wider variety of possible answers.



\section{Experimental Evaluation}
In this section, we assess the effectiveness of various deep-learning models for the tasks outlined in Section~\ref{sec:Task_Description}, utilizing the SimpsonsVQA dataset.

\subsubfour{Baselines:} 
To benchmark SimpsonsVQA, we employed several VQA models,
\textbf{LSTM Q + I~ \cite{Antol_2015_ICCV}}, 
\textbf{MLB~\cite{kim2016hadamard}},
\textbf{MLB+Att~\cite{kim2016hadamard}},
\textbf{MUTAN~\cite{ben2017mutan}}, \textbf{MUTAN+Att~\cite{ben2017mutan}}, \textbf{BUTD \cite{anderson2018bottom}} and \textbf{MCAN \cite{yu2019deep}}, which were evaluated across three tasks with minor adaptations.
For the Answer Correctness task, involving image, question, and answer inputs, we passed the answer through a word-embedding and dense layer, then merged it with the question embedding via element-wise multiplication. 
Additionally, we used recent advanced LVLMs like \textbf{LLava 1.5} \cite{liu2024improved}, \textbf{LLava-Next} \cite{liu2024llavanext}, and \textbf{ChatGpt-4o} \cite{chatgpt4o} in a zero-shot setting. 
For the Conventional VQA task, we fine-tuned LVLMs such as \textbf{ViLT \cite{kim2021vilt}}, \textbf{OFA \cite{wang2022ofa}}, and \textbf{X-VLM \cite{zeng2022x}}. 
For the Question Relevance task, we included  \textbf{QC Simility \cite{ray2016question}} and
\textbf{QQ' Simility \cite{ray2016question}}.

\subsubfour{Metrics:} 
We employ the standard accuracy metric as our primary evaluation criterion. Additionally, we use Precision, Recall, F1-score, and AUC score to ensure a thorough and comprehensive assessment. 

\subsubfour{Implementation details:} All models were implemented according to their original implementation using PyTorch and were trained on a Linux Ubuntu 18.04.1 LTS Dual Intel(R) Xeon(R) Silver CPU @2.20GHz with a GPU NVIDIA Tesla V100. 
Also, all models were trained using the ADAM optimizer~\cite{kingma2014adam} for 15 epochs, while fine-tuned LVLMs were trained for an additional 4 epochs.









\subsection{Results of the Conventional VQA Task}

\setlength{\columnsep}{5pt} 
\setlength{\intextsep}{3pt} 
\setlength{\abovecaptionskip}{1pt} 
\begin{wraptable}{r}{0.2\textwidth}
    \centering
        \caption{Data for the Conventional VQA Task.}
    \label{tab:task_1_dataset}
    \rowcolors{2}{gray!15}{white}

\resizebox{0.19\textwidth}{!}{%
    \begin{tabular}{|ccc|}
    \hline
    \rowcolor{gray!50}

     \textbf{Dataset}& \textbf{\#Images}&\textbf{\#QA Pairs} \\\specialrule{0.1pt}{0pt}{0pt}
     Train & 13,936&60,643\\
     \hline
     Validation &3,451& 9,764  \\
     \hline
     Test &5,409&  10,507  \\
     \hline
     \hline
     Total & 22,796&80,914\\
     \hline
\end{tabular}
}
\end{wraptable}
\noindent
\textbf{SimpsonsVQA dataset:} 
We curated a dataset that includes only triples for which at least 2 workers have assessed as ``Correct'', ensuring the inclusion of only high-quality triples for evaluation.
Table \ref{tab:task_1_dataset} displays the size of the resulting dataset.

\subsubfour{Results analysis:}
Table~\ref{tab:traditional_vqa} presents the performance obtained on the Conventional VQA Task. The results show that traditional VQA models, such as \textbf{LSTM Q + I}, \textbf{SAN}, \textbf{MLB}, and their attention-enhanced variants, exhibit moderate accuracy with particular strength in the ``Yes/No" questions but notably weaker performance in the ``Number" and ``Other" categories.

In contrast, fine-tuned LVLMs such as \textbf{ViLT}, \textbf{X-VLM}, and \textbf{OFA} demonstrate superior performance across all categories, showing a substantial increase in accuracy for ``Number" questions, with \textbf{OFA} achieving an impressive 85.45\% and leading in the ``Other" category at 76.22\%. The overall accuracy rates of these LVLMs significantly exceed those of traditional models, with \textbf{OFA} reaching the highest overall accuracy of 82.00\%. Zero-shot models, including \textbf{LLaVa 1.5}, \textbf{LLaVa 1.6}, and \textbf{ChatGPT-4o}, show inconsistent performance and fail to reach the accuracy levels of fine-tuned models. Specifically, \textbf{ChatGPT-4o}, one of the best LVLMs, achieves an overall accuracy of only 68.32\%, highlighting their limitations on SimpsonVQA. 

\begin{table}[t]
\centering
\caption{Performance on the Conventional VQA Task.}
\label{tab:traditional_vqa}
 \rowcolors{2}{gray!15}{white}
\resizebox{0.32\textwidth}{!}{%
\begin{tabular}{|l|l|cccc|}
\hline

\rowcolor[HTML]{C0C0C0}
\cellcolor[HTML]{C0C0C0}& \cellcolor[HTML]{C0C0C0}& \multicolumn{4}{c|}{\cellcolor[HTML]{C0C0C0}Accuracy} \\ \cline{3-6} 
\rowcolor[HTML]{C0C0C0} 
\multirow{-2}{*}{\cellcolor[HTML]{C0C0C0}ID} & \multirow{-2}{*}{\cellcolor[HTML]{C0C0C0}Model} & Number& Yes/No& Other& All\\ \hline
1 & LSTM Q + I & 0.6102 & 0.8627 & 0.4561 & 0.5823 \\ \hline
2 & SAN & 0.691 & 0.9103 & 0.5336 & 0.6543 \\ \hline
3 & MLB & 0.6042 & 0.8841 & 0.4376 & 0.5786 \\ \hline
4 & MLB+Att & 0.6927 & 0.9038 & 0.6009 & 0.6937 \\ \hline
5 & Mutan & 0.6176 & 0.8962 & 0.4472 & 0.5828 \\ \hline
6 & Mutan+Att & 0.7027 & 0.9262 & 0.6139 & 0.7086 \\ \hline
7 & BUTD & 0.6901 & \textbf{0.9362} & 0.6023 & 0.7037 \\ \hline
8 & MCAN & 0.7347 & 0.9284 & 0.6355 & 0.7209 \\ \hline
\hline
9 & LXMERT & 0.7659 & 0.9149 & 0.6369 & 0.7248 \\ \hline
10 & ViLT & 0.8447 & 0.9101 & 0.7155 & 0.7720 \\ \hline
11 & XVLM & 0.8241 & 0.9215 & 0.7449 & 0.8020 \\ \hline
12 & OFA & \textbf{0.8545} & 0.9335 & \textbf{0.7622} & \textbf{0.8200} \\ \hline
\hline
13 & LLaVa-1.5-LLama-7b & 0.8193 & 0.8241 & 0.4953 & 0.6299 \\ \hline
14 & LLaVa-1.6-Mistral-7b & 0.8279 & 0.8398 & 0.5657 & 0.6784 \\ \hline
15 & GPT-4o-2024-05-13 & 0.8404 & 0.8493 & 0.5789 & 0.6832 \\ \hline
\end{tabular}%
}
\captionsetup{justification=centering, singlelinecheck=false}
\vspace{-0.5cm}
\end{table}

\subsection{Results of the Question Relevance Task}

\subsubfour{SimpsonsVQA dataset:} 
We curated a dataset that comprises images and questions labeled as ``\textit{relevant}'' or ``\textit{irrelevant}'', determined through the majority decision of the workers. 
Table \ref{tab:task_2_dataset}, presents details of this dataset.

\begin{table}[h]
    \centering
        \caption{Data for the Question Relevant task.}
    \label{tab:task_2_dataset}
    \rowcolors{2}{gray!15}{white}

\resizebox{0.4\textwidth}{!}{%
    \begin{tabular}{|ccc||c|}
    \hline
    \rowcolor{gray!50}

     \textbf{Dataset}&\textbf{\#Relevant QA Pairs} & \textbf{\#Irrelevant QA Pairs} & \textbf{Total} \\
     Train & 80,137& 35,526& 115,663\\
     \hline
     Validation & 13,240& 8,709&21,949\\
     \hline
     Tests & 14,680 & 14,241 &28,921\\
     \hline\hline
     Total& 108,057& 58,476 &166,533\\
     \hline

\end{tabular}
}
\end{table}

\subsubfour{Results analysis:} 
Table \ref{tab:question_relevance} provides an overview of the performance of baseline models on the Question Relevance Task.
Notably, all models surpass the accuracy of the majority vote classifier.
Among the assessed models, \textbf{Mutan+Att} stands out, achieving the highest overall accuracy of 87.77\%, significantly benefiting from the addition of attention mechanisms which enhance its capability to discern question relevance. Other traditional VQA models, such as \textbf{QC Similarity} and \textbf{QQ' Similarity}, also perform competently but do not match the effectiveness of \textbf{Mutan+Att}. The integration of attention mechanisms generally boosts model performance, as seen with the improved metrics of \textbf{MLB+Att} over its base version, indicating that attention-augmented models handle the complexities of question relevance tasks more adeptly.

Meanwhile, three zero-shot LVLMs do not outperform the top traditional models because the zero-shot setting inherently lacks task-specific fine-tuning. Among the zero-shot models, \textbf{ChatGPT-4o} achieved the highest accuracy at 83.94\%. Despite this, they still exhibit considerable accuracy, particularly in identifying irrelevant questions.

\begin{table}[t]
\centering
\caption{Performance on the Question Relevance Task.}
\label{tab:question_relevance}
\rowcolors{2}{gray!15}{white}
\resizebox{0.47\textwidth}{!}{%
\begin{tabular}{|l|l|c|cc|cc|cc|}
\hline
\rowcolor[HTML]{C0C0C0} 
\cellcolor[HTML]{C0C0C0} & \cellcolor[HTML]{C0C0C0} & \cellcolor[HTML]{C0C0C0} & \multicolumn{2}{c|}{\cellcolor[HTML]{C0C0C0}Precision} & \multicolumn{2}{c|}{\cellcolor[HTML]{C0C0C0}Recall} & \multicolumn{2}{c}{\cellcolor[HTML]{C0C0C0}F1-Score} \\ \cline{4-9} 
\rowcolor[HTML]{C0C0C0} 
\multirow{-2}{*}{\cellcolor[HTML]{C0C0C0}ID} & \multirow{-2}{*}{\cellcolor[HTML]{C0C0C0}Model} & \multirow{-2}{*}{\cellcolor[HTML]{C0C0C0}Accuracy} & Rel & Irrel & Rel & Irrel & Rel & Irrel \\ \hline
1 & LSTM Q + I & 0.8683 & 0.8501 & 0.8895 & 0.8991 & 0.8365 & 0.8739 & 0.8622 \\
\hline
2 & SAN & 0.8687 & 0.8576 & 0.8809 & 0.8888 & 0.8479 & 0.8729 & 0.8641 \\
\hline
3 & MLB & 0.8686 & 0.8558 & 0.8830 & 0.8914 & 0.8452 & 0.8732 & 0.8637 \\
\hline
4 & MLB+Att & 0.8713 & 0.8545 & 0.8906 & \textbf{0.8997} & 0.8421 & 0.8765 & 0.8657 \\
\hline
5 & Mutan & 0.8650 & 0.8463 & 0.8869 & 0.8970 & 0.8321 & 0.8709 & 0.8586 \\
\hline
6 & Mutan+Att & \textbf{0.8777} & \textbf{0.8770} & 0.8785 & 0.8830 & \textbf{0.8723} & \textbf{0.8800} & \textbf{0.8754} \\
\hline
7 & QC Similarity & 0.8592 & 0.8563 & 0.8628 & 0.8684 & 0.8502 & 0.8623 & 0.8564 \\
\hline
8 & QQ' Similarity & 0.8595 & 0.8553 & 0.8646 & 0.8705 & 0.8487 & 0.8629 & 0.8566 \\
\hline \hline
9 & LLaVa-1.5-LLama-7b & 0.6569 & 0.4148 & \textbf{0.9065} & 0.6030 & 0.8218 & 0.5513 & 0.7243 \\
\hline
10 & LLaVa-1.6-Mistral-7b & 0.7253 & 0.6333 & 0.8201 & 0.7843 & 0.6862 & 0.7008 & 0.7472 \\
\hline
11 & ChatGPT-4o & 0.8394 & 0.8391 & 0.8609 & 0.8702 & 0.8281 & 0.8441 & 0.8493 \\
\hline \hline
12 & Majority Vote & 0.5075 & 0.5075 & 0.0000 & 1.0000 & 0.0000 & 0.6733 & 0.0000 \\
 \hline
\end{tabular}%
}
\captionsetup{justification=centering, singlelinecheck=false}
\vspace{-0.5cm}
\end{table}

\subsection{Results of the Answer Correctness Task}


\subsubfour{SimpsonsVQA dataset:} 
We constructed the dataset using image-question pairs that were deemed relevant as  follows:
When there was unanimous consensus among two or more workers, the majority perspective was assigned as the label for the triple.
If unanimous agreement is not reached, we assign the label ``Ambiguous''.
Table~\ref{tab:task_3_dataset} provides its details.
\vspace{0.2cm}

\begin{table}[h]
    \centering
        \caption{Data for the Answer Correctness task.  C: ``Correct'', AM: ``Ambiguous'', and IC: ``Incorrect''.}
    \label{tab:task_3_dataset}
    \rowcolors{2}{gray!15}{white}
\resizebox{0.47\textwidth}{!}{%
    \begin{tabular}{|c||c||ccc||c|}
    \hline
    \rowcolor{gray!50}

     \textbf{Dataset}& \textbf{\#images}&\textbf{\#C QA Pairs} & \textbf{\#AM QA Pairs} & \textbf{\#IC QA Pairs} & \textbf{Total}\\
     \hline
     Train & 13,961&60,643 & 6,695& 12,799& 80,137\\
     \hline
     Validation & 3,490& 9,764& 1,158 &  2,318 & 13,240\\
     \hline
     Test & 10,507  &5,800 &1,253 & 2,920 & 14,680 \\ 
     \hline
     \hline
     Total & 23,251&80,914& 9,106& 18,037&108,057\\
     \hline

\end{tabular}
}
\end{table}

\subsubfour{Results analysis:}
Table~\ref{tab:answer_correctness} provides a summary of the performance of the baseline models. Given the predominance of “Correct” triples in the data, models generally achieve high performance in this category.
Conversely, performance for the “Ambiguous” category is notably low. The “Incorrect” category shows significant fluctuations, with performance ranging from 17.02\% to 40.26\%.
\textbf{ MLB+Att} and \textbf{Mutan+Att }excel in classifying incorrect triples, likely due to their attention mechanisms and architectures.
Zero-shot large vision language models demonstrate a clear trend of improved performance compared to traditional VQA models.
Especially, \textbf{LLaVa-1.6-Mistral-7b} performs well in the incorrect category. \textbf{ChatGPT-4o} shows balanced performance across all categories, highlighting significant advancements in zero-shot models for assessing the answer correctness.

\begin{table}[t]
\centering
\caption{Performance on the Answer Correctness Task.}
\label{tab:answer_correctness}
 \rowcolors{2}{gray!15}{white}
\resizebox{0.48\textwidth}{!}{%
\begin{tabular}{|c|l|c|ccc|ccc|ccc|}
\hline
\rowcolor[HTML]{C0C0C0} 
\cellcolor[HTML]{C0C0C0} & \cellcolor[HTML]{C0C0C0} & \cellcolor[HTML]{C0C0C0} & \multicolumn{3}{c|}{\cellcolor[HTML]{C0C0C0}Precision} & \multicolumn{3}{c|}{\cellcolor[HTML]{C0C0C0}Recall} & \multicolumn{3}{c|}{\cellcolor[HTML]{C0C0C0}F1-Score} \\ \cline{4-12} 
\rowcolor[HTML]{C0C0C0} 
\multirow{-2}{*}{\cellcolor[HTML]{C0C0C0}ID} & \multirow{-2}{*}{\cellcolor[HTML]{C0C0C0}Model} & \multirow{-2}{*}{\cellcolor[HTML]{C0C0C0}Accuracy} & C & \cellcolor[HTML]{C0C0C0}AM & IC & C & \cellcolor[HTML]{C0C0C0}AM & IC & C & \cellcolor[HTML]{C0C0C0}AM & IC \\ \hline
1 & LSTM Q + I & 0.7660 & 0.7835 & 0.0000 & 0.5107 & \textbf{0.9660} & 0.0000 & 0.1907 & 0.8652 & 0.0000 & 0.2777 \\ 
\hline
2 & SAN & \textbf{0.7712} & 0.8023 & 0.0250 & 0.5172 & 0.9476 & 0.0001 & 0.3067 & 0.8689 & 0.0018 & 0.3850 \\ 
\hline
3 & MLB & 0.7672 & 0.7750 & 0.0000 & 0.5604 & 0.9735 & 0.0000 & 0.1190 & 0.8669 & 0.0000 & 0.1963 \\ 
\hline
4 & MLB+Att & 0.7530 & 0.8119 & 0.1753 & 0.4736 & 0.9078 & 0.0663 & 0.3465 & 0.8571 & 0.0960 & 0.4001 \\ 
\hline
5 & Mutan & 0.7742 & 0.7969 & 0.1570 & 0.5459 & 0.9586 & 0.0051 & 0.2691 & \textbf{0.8703} & 0.0009 & 0.3904 \\ 
\hline
6 & Mutan+Att & 0.7500 & 0.8180 & \textbf{0.1909} & 0.4601 & 0.8968 & 0.0819 & 0.3710 & 0.8556 & 0.1139 & 0.4107 \\
\hline \hline
7 & LLaVa-1.5-LLama-7b & 0.6933 & \textbf{0.9666} & 0.0167 & 0.0571 & 0.7211 & 0.0972 & 0.4394 & 0.8260 & 0.0285 & 0.1012 \\
\hline
8 & LLaVa-1.6-Mistral-7b & 0.7431 & 0.8550 & 0.0247 & 0.6486 & 0.8473 & \textbf{0.2719} & \textbf{0.4778} & 0.8512 & 0.0453 & \textbf{0.5502} \\ 
\hline
9 & ChatGPT-4o & 0.7198 & 0.7947 & 0.1761 & \textbf{0.6802} & 0.8912 & 0.2269 & 0.4574 & 0.8402 & \textbf{0.1983} & 0.5470 \\ 
\hline
\end{tabular}%
}
\captionsetup{justification=centering, singlelinecheck=false}
\vspace{-0.5cm}
\end{table}

\vspace{-0.12cm}
\subsection{Discussion}
Drawing from our experimental findings, several key observations come to light. 
First, among the three VQA tasks mentioned earlier, the assessed baseline models demonstrate effectiveness in the Conventional VQA Task, while facing challenges in predicting questions that do not fall into the categories of ``yes/no'' or ``number'' questions. 
This outcome is comprehensible given that these models are crafted with a distinct emphasis on the conventional VQA task.
Second, in the Question Relevance task, existing models showed comparable performance.
A key challenge is determining the validity of a question based on the visual content of the image.
This requires models to understand the question and assess its appropriateness within the context of the image.
The Answer Correctness task presents a significant challenge, more complex than traditional VQA tasks.
In addition to understanding the image and question to generate answers, models must evaluate the interplay between the image, question, and answer to classify responses as ``correct'', ``incorrect'', or ``ambiguous''.
This task is a more advanced version of the standard VQA exercise, requiring a deeper, more nuanced level of understanding.

\section{Potential Negative Societal Impact}
While the SimpsonsVQA holds valuable educational and learning applications, there are several potential negative implications that need to be acknowledged.

\subsubsubfour{Stereotyping and Bias:} 
The dataset, derived from The Simpsons TV show, may contain stereotypes, biases, or cultural references that could perpetuate negative perceptions. Using it in educational settings risks learners absorbing these biases or incorrect information.


\subsubsubfour{Cognitive and Emotional Impact:} AI-generated content in education can affect learners' cognitive and emotional development. Ensuring the content is age-appropriate, respectful, and conducive to positive learning is crucial.


\subsubsubfour{Over-reliance on AI-driven tools:} Excessive dependence on AI in education may diminish the role of educators and human interaction. Hence, it's important to balance technological advancements with human guidance.



\vspace{-0.15cm}

\section{Conclusion \& Future Work}
\textcolor{black}{
In conclusion, our findings show that advanced LVLMs like LLaVa and ChatGPT face significant challenges when processing cartoon images, such as those in the SimpsonsVQA dataset, due to their training on photorealistic data. SimpsonsVQA provides a valuable resource for evaluating and fine-tuning these models, enhancing their robustness in handling cartoon-based tasks and pushing the limits of VQA capabilities.
}

\textcolor{black}{By leveraging The Simpsons' consistent visual style, this dataset contributes to the development of intelligent systems for educational applications, promoting inquiry-based learning. However, we acknowledge that the automatically generated irrelevant questions and incorrect answers may introduce a domain gap compared to human learners' errors. To address this, we plan a human study to ensure better alignment with real learner behaviors. Additionally, we recognize that different cartoon styles may create distinct image domains, which will be explored in future work to improve model adaptability and expand the dataset's value for broader visual applications.}

\vspace{0.15cm}

\hrule
\vspace{0.15cm}

{

\subsubsubfour{Licensing:} The dataset is licensed under a Creative Commons Attribution-NonCommercial-ShareAlike 4.0 International (CC BY-NC-SA 4.0)~\footnote{\scriptsize \url{https://creativecommons.org/licenses/by-nc-sa/4.0/}}. 


\subsubsubfour{Ethical considerations:} In creating SimpsonsVQA, there was no collection or publication of any personal or critical data related to the AMT workers. 
The annotators responsible for labeling the SimpsonsVQA dataset were compensated fairly for their efforts, adhering to the minimum wage standards set by the platform.
}

{\small
\bibliographystyle{ieee_fullname}
\balance
\bibliography{biblio}
}

\end{document}